\documentclass[sigconf]{acmart}

\AtBeginDocument{%
  \providecommand\BibTeX{{%
    \normalfont B\kern-0.5em{\scshape i\kern-0.25em b}\kern-0.8em\TeX}}}

\usepackage{booktabs}       
\usepackage{amsfonts}       
\usepackage{nicefrac}       
\usepackage{microtype}      
\usepackage{multirow, makecell}
\usepackage{kotex}
\usepackage{xcolor}
\usepackage{xspace}
\usepackage{enumitem}
\usepackage{amsmath}
\usepackage{mathtools}
\usepackage{nameref}
\usepackage{algorithm}
\usepackage{balance}

\DeclareMathOperator*{\argmin}{arg\,min}

\newcommand{\PreserveBackslash}[1]{\let\temp=\\#1\let\\=\temp}
\newcolumntype{C}[1]{>{\PreserveBackslash\centering}p{#1}}

\definecolor{Gray}{gray}{0.95}

\copyrightyear{2024}
\acmYear{2024}
\setcopyright{rightsretained}
\acmConference[MM '24]{Proceedings of the 32nd ACM International Conference on Multimedia}{October 28-November 1, 2024}{Melbourne, VIC, Australia}
\acmBooktitle{Proceedings of the 32nd ACM International Conference on Multimedia (MM '24), October 28-November 1, 2024, Melbourne, VIC, Australia}
\acmDOI{10.1145/3664647.3680955}
\acmISBN{979-8-4007-0686-8/24/10}

\begin{document}

\title[Training Spatial-Frequency Visual Prompts and Probabilistic Clusters
for Black-Box Transfer Learning]{Training Spatial-Frequency Visual Prompts and Probabilistic Clusters
for Accurate Black-Box Transfer Learning}



\author{Wonwoo Cho}
\authornote{Co-first authors.}
\author{Kangyeol Kim}
\authornotemark[1]
\affiliation{%
  \institution{Korea Advanced Institute of Science and Technology}
  \city{Daejeon}
  \country{Republic of Korea}
}
\affiliation{%
  \institution{Letsur Inc.}
  \city{Seoul}
  \country{Republic of Korea}
}
\email{{wcho, kangyeolk}@kaist.ac.kr}

\author{Saemee Choi}
\authornotemark[1]
\affiliation{%
  \institution{Korea Advanced Institute of Science and Technology}
  \city{Daejeon}
  \country{Republic of Korea}
}
\email{saemee99@kaist.ac.kr}

\author{Jaegul Choo}
\affiliation{%
  \institution{Korea Advanced Institute of Science and Technology}
  \city{Daejeon}
  \country{Republic of Korea}
}
\affiliation{%
  \institution{Letsur Inc.}
  \city{Seoul}
  \country{Republic of Korea}
}
\email{jchoo@kaist.ac.kr}



\begin{abstract}
   {Despite the growing prevalence of black-box pre-trained models (PTMs) such as prediction API services, there remains a significant challenge in directly applying general models to real-world scenarios due to the data distribution gap. Considering a data deficiency and constrained computational resource scenario, this paper proposes a novel parameter-efficient transfer learning framework for vision recognition models in the black-box setting. Our framework incorporates two novel training techniques. First, we align the \emph{input space} (\emph{i.e.}, image) of PTMs to the target data distribution by generating visual prompts of spatial and frequency domain. Along with the novel spatial-frequency hybrid visual prompter, we design a novel training technique based on probabilistic clusters, which can enhance class separation in the \emph{output space} (\emph{i.e.}, prediction probabilities). In experiments, our model demonstrates superior performance in a few-shot transfer learning setting across extensive visual recognition datasets, surpassing state-of-the-art baselines. Additionally, we show that the proposed method efficiently reduces computational costs for training and inference phases.}
\end{abstract}





\maketitle


\section{Introduction}


In recent years, machine learning (ML) models pre-trained with large-scale data have shown significant performance improvements in visual recognition, even in zero-shot classification~\cite{brown2020language, rombach2022high, radford2021learning}. However, it is difficult to properly generalize to distinct data distributions, necessitating the development of \textit{domain-specific} ML models. Although transfer learning has been widely investigated to re-purpose pre-trained models (PTMs) by transferring the knowledge learned in a source-domain task to a target domain, it is expensive to obtain well-organized target-domain datasets (\emph{e.g.,} medical datasets). In such data-deficient scenarios, parameter-efficient transfer learning (PETL)~\cite{tian2020rethinking, houlsby2019parameter, zhang2021tip, zhou2022learning}, which aims to adapt classifiers to downstream tasks by tuning only a small number of additional parameters, can be an economical and effective solution.

In addition to the data deficiency issue, the implementation of transfer learning in real-world industrial scenarios poses significant challenges. Following the great success of ChatGPT~\footnote{\url{https://chatgpt.com/}}, there has been a noticeable trend towards the release of access-restricted PTMs in the form of APIs. Essentially, the intricate details of recent enterprise-level ML models, including their architecture, weights, training data, and techniques, are often kept confidential. While such API services offer efficient solutions for users having limited resources—such as small companies or individuals who face computational limitations when computing gradients through large-scale ML models—it remains challenging to adapt these opaque, \emph{black-box} PTMs to domain-specific tasks that require tailored applications.

To address this challenge, we propose a novel tuning methodology for visual recognition models in a black-box environment. In this setting, the method utilizes \emph{only the available input-output responses} (input data items and their prediction probabilities) for tuning. Given that the input-output response represents the only accessible information, one may strategically optimize the input space to effectively enhance model performance within such constraints. Although it may be feasible to tune the output space in white-box scenarios (\emph{e.g.}, by applying a linear layer to the penultimate layer), these methods can become intractable in black-box settings. In this context, one can primarily focus on the concept of \emph{visual prompting}~\cite{bahng2020learning, unleashing22}, which aims to learn pixel-level perturbations adjusting the input space to better match the target distribution.


To date, studies focusing on visual prompting in the black-box setting, where all parameters and gradients are not directly accessible, are comparatively limited when contrasted with the white-box setting.
Notably, only a few investigations have explored this area.
The black-box adversarial reprogramming (BAR) method~\cite{reprogramming20}, marks a pioneering effort in emphasizing the importance of input-space adaptation for black-box PTMs.
This method employs frame-shaped visual prompts (VPs) for effective adaptation.
In a subsequent development, BlackVIP~\cite{blackvip23} introduces input-dependent VPs through the use of an auxiliary encoder, which is specifically designed to extract features from images.
This approach underscores the versatility and adaptability of visual prompting, significantly enhancing compatibility of models within specialized domains.


Building upon existing frameworks, our research aims to enhance the robustness of black-box PETL by designing \emph{1) effective visual prompting} and \emph{2) novel output-space tuning} strategies.
First, we design a \emph{spatial-frequency hybrid visual prompter} with the primary goal of effectively manipulating images in the frequency domain.
Techniques such as low-frequency perturbations, which maintain the structural integrity of images, have shown promising results in studies related to domain adaptation~\cite{yang2020fda, yang2020phase, xu2021fourier, wang2023fvp} and adversarial attacks~\cite{guo2018low, jia2022exploring}.
However, the efficacy of these frequency-domain manipulations can vary across different datasets.
Thus, we also integrate spatial-domain VPs by designing a spatial-frequency hybrid visual prompter, where this integration is carefully managed to avoid conflicts between the two domains, resulting in an effective adaptation strategy for PTMs across various target datasets.

The motivation for our second component, output-space tuning, comes from the observation that prediction probabilities of PTMs often show minimal differences across classes when applied to a new domain.
This similarity complicates the extraction of a distinct and robust learning signal, posing a challenge for effective domain adaptation.
In white-box PETL, two baseline approaches to address this issue are linear probing for visual encoders and text prompt tuning for vision-language models.
However, the restrictive nature of the black-box setting makes it challenging to modify the output space directly.
As a workaround, we propose trainable \emph{cluster-based prediction refinement} to enhance the
class-distinguishability in the output space.
This method involves using auxiliary simplex prototypes to disentangle and reorganize the output space, making it more structured and easier to differentiate among classes.


In our experiments, we rigorously assess the performance of our model through few-shot transfer learning across a wide range of benchmark datasets, thereby showcasing the robustness and effectiveness of our proposed methodologies.
Additionally, we conduct a thorough analysis of each component of our method to provide evidence supporting its architectural design.
Our experiments further include an examination of computational overheads, specifically targeting memory efficiency and the reduction in training time, to highlight the practical advantages of our approach.

\section{Related Work}
\label{related_work}

\textbf{Visual prompting.}\quad
The recent advent of foundation PTMs,
as highlighted in key studies~\cite{brown2020language,rombach2022high,radford2021learning},
has significantly increased interest in deploying these models for real-world applications.
Despite their potential, a notable performance drop is often observed when these general-purpose models are applied directly to specialized domains.
To bridge this gap, the concept of PETL~\cite{tian2020rethinking,zhang2021tip,zhou2022learning,jia2022visual} has emerged as a prominent research direction. 

Visual prompting, one of promising approaches of PETL, aims to adapt the input distribution for a specific target model by introducing learnable, \emph{dataset-specific perturbations at the pixel level}~\cite{exploring22, adversarialrobustness23}.
Following this initial approach, subsequent studies have concentrated on refining the design of VPs.
For example, Wu \emph{et al.}~\cite{unleashing22} introduces a novel approach for integrating an image with a VP, creating a synergistic blend that enhances model performance.
Additionally, Huang \emph{et al.}~\cite{diversitymeta23} advocates for the development of multiple group-specific VPs. This method is designed to address the complex issue of distribution shift, aiming to more closely align with the original data distribution of the target model.


\vspace{1\baselineskip}
\noindent\textbf{Black-box Visual Prompting.}\quad
Recently, a visual prompting in a black-box setting has been studied~\cite{reprogramming20,blackvip23} with an increasing attention towards privacy models.
To the best of our knowledge, BAR~\cite{reprogramming20} represents the initial effort to incorporate VP into a black-box setting. 
This approach aims to effectively repurpose models pre-trained on natural images for applications in the medical domain.
BlackVIP~\cite{blackvip23} propose an input-dependant VP design and the corresponding optimization method for black-box setting, and show a promising results on various benchmark datasets.

However, there exist some limitations in the early studies. For instance,
BAR's frame-like VP does not effectively manipulate input images, leading to ineffectiveness in some cases.
BlackVIP uses an input-dependent full-resolution VP, whose performance can be limited for some datasets like SVHN.
Since frequency-domain components can play a crucial role in such cases, as observed in~\cite{autovp23}, we develop a hybrid visual prompting method to address these issues.
Several studies~\cite{testtimefrequency23,explicitvisualprompt23,fouriervisualprompt23} have ventured into examining the impact of VPs within the frequency domain.
Initial investigations~\cite{explicitvisualprompt23,fouriervisualprompt23} have shown that applying visual prompting techniques to the frequency domain can significantly enhance performance across a broad spectrum of downstream tasks.
These tasks include unsupervised domain adaptation for segmentation, presenting the versatility and effectiveness of frequency-based visual prompting.


\begin{figure*}[t] 
\begin{center} 
\includegraphics[width=1\linewidth]{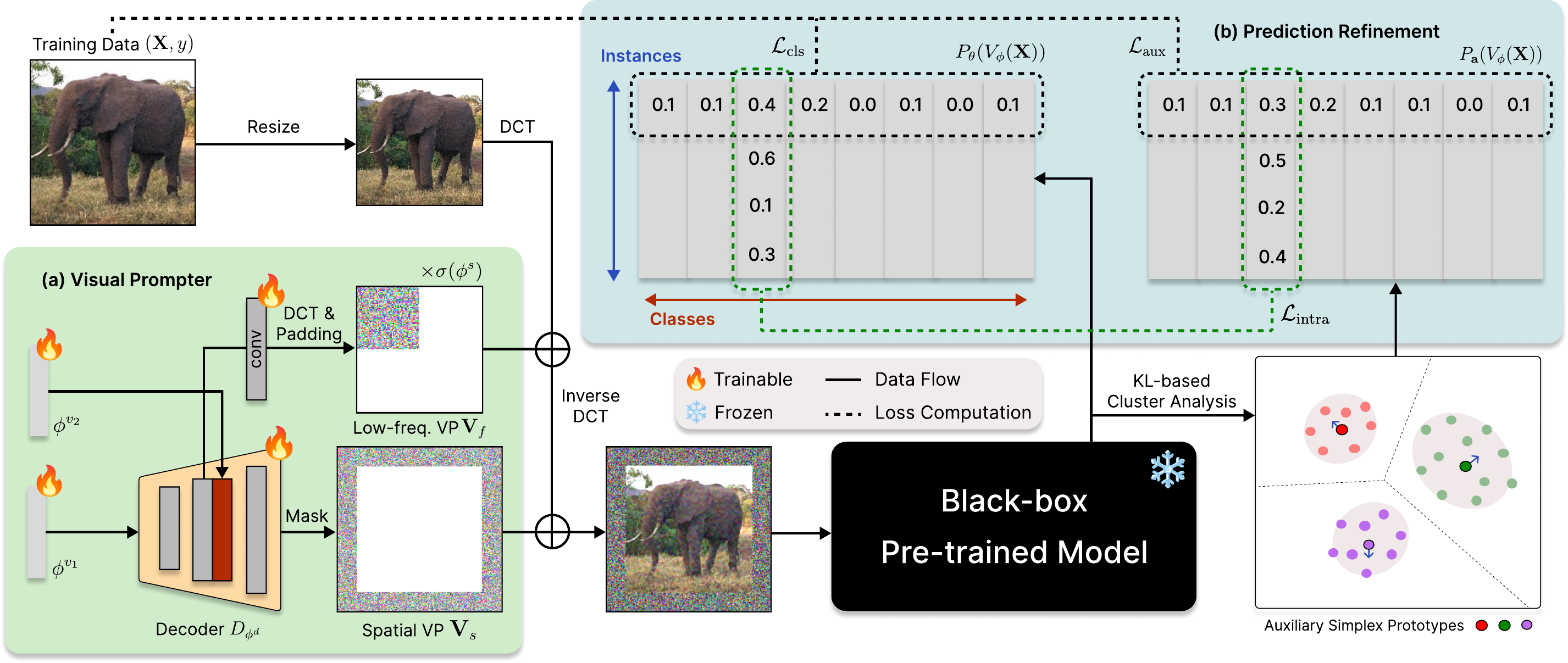}
\end{center}
\caption{A overall training workflow of our proposed method.
(a) Our visual prompter consists of
a single decoder $D_{\phi^d}$ and two trigger vectors ($\phi^{v_1}, \phi^{v_2}$), where the decoder simultaneously generates two VPs in spatial and frequency domains, respectively.
Also, the learnable scaling parameter $\phi^s$
controls the effect of the low-frequency VP according to its efficacy.
(b) After obtaining prediction probabilities
$P_{\theta}(V_{\phi}(\mathbf{X}))$,
we conduct prediction refinement via KL-based cluster analysis.
During training, we utilize auxiliary simplex prototypes to
enhance the effectiveness of clustering based prediction refinement.
}
\label{fig:overview}
\end{figure*}

\vspace{1\baselineskip}
\noindent\textbf{Output space alignment.}\quad
To better align input data with output predictions in white-box scenarios, techniques such as adding a linear layer at the end of the network (\emph{i.e.}, linear probing) can be employed. In the field of vision-language models, efforts to align the output space often involve tuning text embeddings~\cite{radford2021learning, li2022blip, li2023blip, li2021align}. Research on text prompt tuning~\cite{zhou2022learning, zhou2022conditional, zang2022unified} serves as foundational inspiration for our cluster-based prediction refinement approach. Both CoOp~\cite{zhou2022learning} and CoCoOp~\cite{zhou2022conditional} introduce methods for adjusting text prompts with learnable components, demonstrating their effectiveness in achieving better alignment between images and text. Additionally, Zhang~\emph{et al.}~\cite{zang2022unified} highlights the critical role of text prompt tuning in reducing low inter-class variance of text features, underscoring the potential impact of such adjustments.

In black-box settings, as described in~\cite{blackvip23}, the only resources available for training are input samples and their corresponding prediction probabilities. As a result, there have been few attempts to tune the output space in these environments. Although some studies~\cite{guo2023black, yu2023black, ouali2023black} have explored black-box tuning of text prompt token embeddings, these methods assume that logits or input embeddings are accessible. In this paper, we provide a thorough comparison with studies relevant to our method in the black-box PETL setting, where only inputs and output probability predictions are available.

\section{Preliminaries}

\noindent\textbf{Problem statement.}\quad
Throughout this paper,
we denote $\mathcal{D} \equiv \{(\mathbf{X}_{i}, y_{i})\}_{i=1}^N$ as a training set comprising $N$ data items, where $\mathbf{X}_{i} \in \mathbb{R}^{h \times w \times c}$ is a $c$-channel 2D image of size $h \times w$, and $y_i \in \{1, \cdots, K\}$ is the corresponding class label.
Given $\mathcal{D}$, this paper aims to construct a visual prompter $V_{\phi}$ to conduct PETL of a pre-trained visual recognition model $P_{\theta}$, as in previous works~\cite{exploring22, unleashing22, diversitymeta23}.

We follow the black-box setting of BAR~\cite{reprogramming20},
which first defines the problem of black-box visual prompting.
Specifically, it is assumed that $P_{\theta}$ is frozen and one can only access \emph{\textbf {1) each input sample}} $(\mathbf{X}_{i},y_i)$ and \emph{\textbf{2) its prediction probabilities}} $P_{\theta}(V_{\phi}(\mathbf{X}))$.
As it is intractable to compute gradient through $V_{\phi}$, the objective
\begin{equation*}
  \phi^* = \argmin_{\phi} \mathbb{E}_{\mathcal{D}\sim (\mathbf{X}, y)} \left[\mathcal{L}(P_{\theta}(V_{\phi}(\mathbf{X})), y)\right]
\end{equation*}
can be optimized by zeroth-order optimization~\cite{liu2020primer} to find an optimal $\phi$, where $\mathcal{L}$ is a task-specific loss function.

\vspace{1\baselineskip}
\noindent\textbf{Visual prompting.}\quad
In BAR~\cite{reprogramming20},
its visual prompter is defined by
$V_{\phi}(\mathbf{X}) = \texttt{Pad}(\mathbf{X}_r) + \mathbf{V}_\phi$,
where $\texttt{Pad}(\mathbf{X}_r)$ is a resized \& zero-padded version of $\mathbf{X}$ and $\mathbf{V}_\phi\in \mathbb{R}^{h \times w \times c}$ is a VP (zero-masked at center)
directly added to $\mathbf{X}_r$, respectively.
These frame-like VPs apply trainable pixel-level perturbations to the periphery of images.
Various designs of frame VPs are possible; for instance, one might downsample an image and pad the VPs outside~\cite{reprogramming20, unleashing22}, or overlap the VPs directly onto the original image~\cite{exploring22}.
Further techniques, such as those considering multiple subdomain-specific visual prompts (VPs) \cite{diversitymeta23} and selecting VP sizes that vary according to dataset characteristics \cite{tsao2023autovp}, have also been explored.

BlackVIP~\cite{blackvip23}, a recently proposed black-box visual prompting method, addresses the challenges associated with the high parameter requirements and limited performance of BAR. The approach effectively resolves these issues through the implementation of an encoder-decoder architecture,
which can be defined by
\begin{equation}\label{eq:blackvip}
    V_{\phi}(\mathbf{X}) =
    \mathbf{X} +  \epsilon \cdot 
    D_{\phi^d}(E(\mathbf{X}) \oplus \phi^v).
\end{equation}
In Eq.~\eqref{eq:blackvip},
$E$ is a self-supervised feature extractor, $D_{\phi^d}$ is a decoder generating a VP of size ($h\times w \times c)$,
$\phi^v$ is a trainable trigger vector,
and $\epsilon \in (0,1]$ is a hyperparameter.

\vspace{1\baselineskip}
\noindent\textbf{Zeroth-order optimization.}\quad
In the white-box scenario, where there is complete access to PTMs, optimizing the loss function and retrieving its gradient through back-propagation are straightforward.
However, in the case of a black-box model, where only the model outputs are available, back-propagation is infeasible due to the unavailability of the gradient.
In such situations, zeroth-order optimization emerges as a viable solution to this challenge. The two principal components of this method are 1)  gradient estimation and 2) gradient descent using the estimated gradient.

In BlackVIP~\cite{blackvip23},
the training parameters $\phi = \{\phi^d, \phi^v\}$ are optimized by the proposed SPSA-GC algorithm,
which is a modified version of the
simultaneous perturbation stochastic approximation (SPSA)~\cite{spsa92}
based on the principle of Nesterov’s accelerated gradient~\cite{nesterov83}.
By using the conventional cross-entropy loss as $\mathcal{L}$, the authors conduct
multi-point gradient estimation for SPSA, \emph{i.e.}, the $t$-th iteration estimated gradient for the current parameters $\phi_t$ is
\begin{equation}\label{eq:spsagrad}
\hat{g}_t(\phi_t) = \frac{1}{S} \sum_{s=1}^S \frac{\mathcal{L}(\phi_t + c_t \Delta_{s}) - \mathcal{L}(\phi_t - c_t \Delta_s)}{2 c_t} \Delta_s^{-1},
\end{equation}
where $c_{i} \in [0,1]$ is a decaying parameter and
$\Delta_{s}$ is a perturbation vector sampled from mean-zero distributions that satisfy finite inverse momentum condition such as Rademacher and segmented uniform distribution.
For more in-depth description of the SPSA-GC optimization, readers are referred to the BlackVIP method~\cite{blackvip23}.

\vspace{1\baselineskip}
\noindent\textbf{Discrete cosine transform.}\quad
The discrete cosine transform (DCT) converts spatial domain data (image pixels) into the frequency domain. Each component of the DCT output represents a specific frequency in the original image, with low-frequency components in the top-left corner of the DCT matrix and high-frequency components towards the bottom-right. This arrangement distinguishes between general trends and finer details in the image. Isolating and modifying these frequencies allows for targeted analysis and manipulation, offering insights into how different image characteristics influence the behavior of pre-trained models.

2D DCT operation on an $n \times n$-sized image can be defined by

\begin{equation}
\begin{split}
D(u,v) = \alpha(&u) \alpha(v) \cdot \\ & \sum_{x=0}^{N-1} \sum_{y=0}^{N-1}  f(x,y)  \cos\left[\frac{(2x+1)u\pi}{2n}\right] \cos\left[\frac{(2y+1)v\pi}{2n}\right],
\end{split}
\end{equation}
where $D(u,v)$ is the DCT coefficient at frequencies 
$u$ and $v$, 
$f(x,y)$ is the pixel intensity at coordinates 
$(x,y)$, and $\alpha(u)$ and $\alpha(v)$ are normalization factors, respectively.

\section{Methods}

\noindent\textbf{Overview.}\quad
To further improve effectiveness when tuning a PTM in a black-box environment,
we propose novel visual prompting and output-space tuning methods,
where our contribution can be summarized as follows:

\begin{enumerate}[itemsep=2pt]
  \item[$\bullet$]
  For effective tuning of the input space of $P_{\theta}$ by using few-shot target data samples, we design a novel visual prompter $V_\phi$, which simultaneously provides prompts in the spatial and frequency domains. (Section~\ref{sec3.1})
  \item[$\bullet$] We propose a cluster-based class prediction strategy to flexibly manipulate the output probabilities of $P_{\theta}$. (Section~\ref{sec3.2})
  \item[$\bullet$] Various techniques are developed to further stabilize the training process of our visual prompting method. (Section~\ref{sec3.3})
\end{enumerate}

\label{eq:spsa-gc}



\subsection{Spatial-Frequency Visual Prompter}\label{sec3.1}


Recall the design of the visual prompter described in Eq.~\eqref{eq:blackvip}.
This setup demonstrates that an encoder-decoder-based VP generation can be more effective than directly tuning pixel-level perturbations in resource-constrained black-box scenarios. The underlying assumption is that a PTM models the data geometry within a target dataset based on its own feature space. Providing different visual prompts based on the geometry of data (e.g., sub-domains or clusters) is likely to be effective. An encoder can be used based on this assumption, applying unique transfer rules for each image.




However, in a black-box scenario where the PTM is not disclosed, using an arbitrary encoder may impose incorrect assumptions about the data geometry.
An inappropriate choice of encoder can lead to unintended outcomes and significant variability in effectiveness across different datasets.
To address the issue, we design an \emph{encoder-free} visual prompter, where we aim to create a novel visual prompter that can effectively adapt the input space of PTMs across various datasets.
Inspired by an observation that manipulation in the low-frequency domain can facilitate efficient learning of VPs in certain datasets~\cite{autovp23}, we consider the following aspects:



\begin{enumerate}[itemsep=2pt]
  \item[$\bullet$] To enhance effectiveness, we incorporate a frequency-domain prompting strategy into our visual prompter.
  \item[$\bullet$] The impact of frequency-domain VPs can vary across datasets; therefore, we make their influence adjustable for each dataset and compensate with spatial-domain VPs.
  \item[$\bullet$] Considering a resource-constrained setting, we generate VPs in both the frequency and spatial domains in a memory-efficient manner, by using a single decoder structure.
\end{enumerate}

\begin{figure*}[t] 
\begin{center} 
\includegraphics[width=0.85\linewidth]{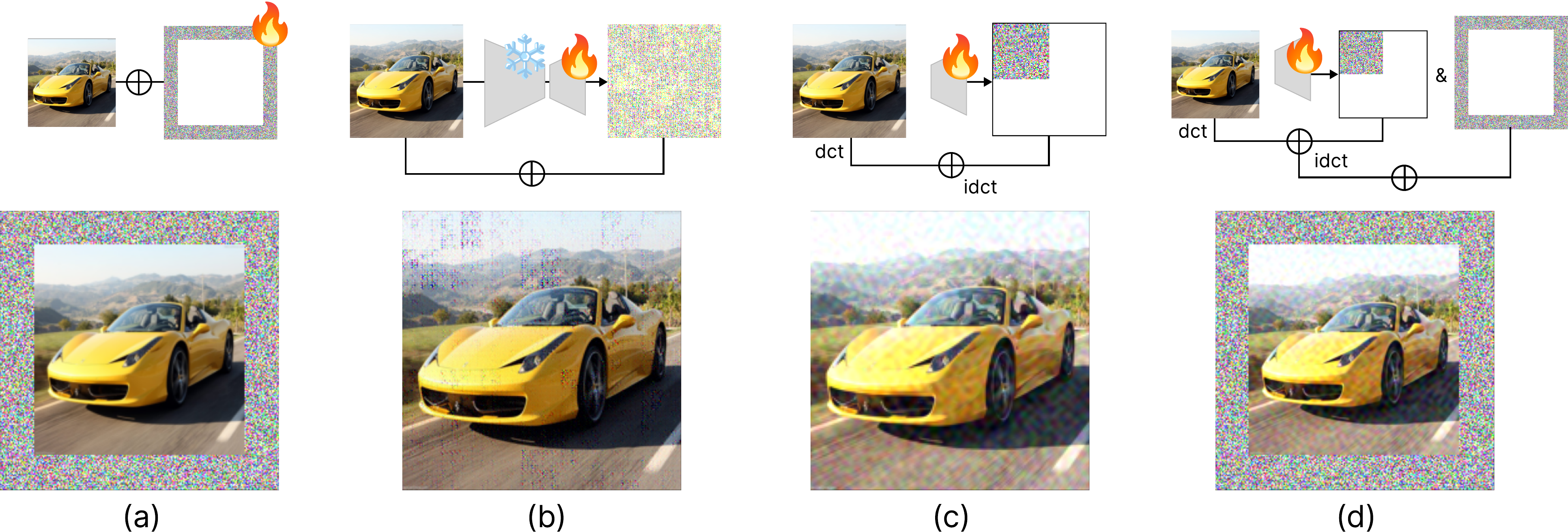}
\end{center}
\caption{An illustration of various visual prompting methods.
(a) Visual prompting in the spatial domain is achieved by padding a VP outside each image. The VP itself can be trained using the principles of BAR~\cite{reprogramming20}.
(b) The spatial-domain visual prompting of BlackVIP~\cite{blackvip23}, where VPs are generated by an encoder-decoder network.
(c) Low-frequency visual prompting, where a decoder
makes low-frequency visual prompts located in
top-left corner (low-frequency in DCT).
(d) Our Spatial-frequency visual prompting method,
where spatial- and frequency-domain VPs are simulteneously constructed by a single decoder.
}
\label{fig:visualprompting}
\end{figure*}

As illustrated in Fig.~\ref{fig:overview}(a).
our visual prompter consists of a single decoder taking a learnable trigger vector $\phi^{v_1}$ as input.
We use additional one $\phi^{v_2}$ to enable flexible generation of a spatial-domain VP.
Denoting spatial and frequency VPs as $\mathbf{V}_s\in \mathbb{R}^{h \times w \times c}$ and $\mathbf{V}_f \in \mathbb{R}^{h_f \times w_f \times c}$, respectively,
our decoder $D_{\phi^d}$ is
\begin{equation}\label{eq:decoder}
    (\mathbf{V}_f, \mathbf{V}_s)= D_{\phi^d} (\phi^{v_1}, \phi^{v_2}),
\end{equation}
where $h \gg h_f$ and $w \gg w_f$.
Based on Eq.~\eqref{eq:decoder},
our visual prompting process using $V_\phi$ can be described as follows:

\vspace{1\baselineskip}
\noindent\textbf{Frequency-domain prompting.}\quad
Image manipulation in the low-frequency band has become a popular strategy in adversarial attacks and adaptations~\cite{guo2018low, jia2022exploring} to minimize redundant noise in the spatial domain and preserve the perceptual similarity of images. The low-frequency region, which accounts for most of an image's energy, is associated with content, whereas the high-frequency region pertains to edge and texture information.

Among the frequency analysis techniques such as fast Fourier transform and discrete wavelet transform,
we employ  discrete cosine transform (\texttt{dct}) and its inversion (\texttt{idct}) in our VPs.
Given $\mathbf{X}$, downsample the image to $\mathbf{X}_r \in \mathbb{R}^{h_d \times w_d \times c}$, where $h_d > h_f$ and $w_d > w_f$.
Then, frequency-domain prompted $\mathbf{X}_r$ is
\begin{equation}\label{eq:freqprompt}
  \mathbf{X}_{r,fvp} = 
  \texttt{idct}(\texttt{dct}(\mathbf{X}_r) + \sigma(\phi^s) \cdot \texttt{dct}(\mathbf{V}_f))),
\end{equation}
where $\sigma$ and $\phi^s$ denote the sigmoid operation and a learnable scalar parameter for scaling the strength of frequency-domain prompting, respectively.
In Eq.~\eqref{eq:freqprompt}, we pad the bottom and right sides of the $\texttt{dct}(\mathbf{V}_f)$ with zeros to match it to the size of $\texttt{dct}(\mathbf{X}_r)$.

\vspace{1\baselineskip}
\noindent\textbf{Spatial-domain prompting.}\quad
After frequency-domain prompting, we apply spatial-domain prompting by
\begin{equation}\label{eq:spatialprompt}
  V_\phi(\mathbf{X}) = \texttt{Pad}(\mathbf{X}_{r,fvp}) + \texttt{Mask}(\mathbf{V}_s, \mathbf{X}_{r}),
\end{equation}
where $\texttt{Mask}(\mathbf{V}_s, \mathbf{X}_{r})$ is an operation masking the center portion of $\mathbf{V}_s$ with zeros, matching the size of $\mathbf{X}_{r}$.
In summary, our visual prompter $V_\phi$
applies both spatial- and frequency-domain VPs to each input image
based on a single encoder $D_\phi^d$, two trigger vectors $(\phi^{v_1}, \phi^{v_2})$, and a scaling parameter $\phi^s$, \emph{i.e.,}
$\phi = \{\phi^d, \phi^{v_1}, \phi^{v_2}, \phi^s\}$.

\vspace{1\baselineskip}
In this section, we design an encoder-free visual prompter that uses a single decoder instead of two separate decoders. This simplifies the learning process, which is crucial given the unstable nature of black-box optimization.
To avoid information overlap between spatial and frequency domains information and complement each other, we implement two strategies. For the frequency-domain prompt, we extract intermediate features from the decoder, apply a 1x1 convolution, and transform these features into the frequency domain using DCT. For the spatial-domain prompt, we introduce an additional learnable trigger vector. These strategies allow us to generate effective prompts for both domains using one decoder, leveraging shared information while reducing conflicts.

\subsection{Cluster-based Prediction Refinement}~\label{sec3.2}



In the output space of a black-box PTM, only prediction probabilities are accessible.
These probabilities are typically computed using class-specific parameters trained with source-domain data. For example, if logits are computed based on distances between class prototypes and output features, each class prototype depends on the source-domain classes. Thus, such prototypes may be inadequate for accurately representing target-domain classes.
Moreover, if the class prototypes remain fixed and are not adjusted for different domains, they may not provide a strong learning signal to the input space. 
To refine potentially erroneous predictions caused by inherent prototypes, we leverage the principle of probabilistic clustering
and then define auxiliary simplex prototypes.




\vspace{1\baselineskip}
\noindent\textbf{Probabilistic clustering for prediction refinement.}\quad
Assuming that the largest index of $P_{\theta}(V_{\phi}(\mathbf{X}))$ can be a sub-optimal class prediction in some cases, one can use clustering methods for prediction refinement.
Since only prediction probabilities are accessible, we apply KL k-means clustering~\cite{chaudhuri2008finding} with $\{P_{\theta}(V_{\phi}(\mathbf{X}_i))\}_{i=1}^N$,
which is a probabilistic clustering method for simplex data.

By clustering $\{P_{\theta}(V_{\phi}(\mathbf{X}_i))\}_{i=1}^N$ into $K$ clusters based on the KL divergence metric, where $K$ is the number of target classes, we obtain class-wise simplex mean vectors, \emph{i.e.}, $\mathbf{m} = \{m_k\}_{k=1}^K$.
Then, a refined version of $P_{\theta}(V_{\phi}(\mathbf{X}))$ can be formulated by
\begin{equation}\label{eq:clusterprob}
    P_\mathbf{m}(V_{\phi}(\mathbf{X}))[k] = \frac{\exp(-\text{KL}(P_{\theta}(V_{\phi}(\mathbf{X}))||m_k))}{\sum_{l=1}^K \exp(-\text{KL}(P_{\theta}(V_{\phi}(\mathbf{X}))||m_l))},
\end{equation}
where $-\text{KL}(P_{\theta}(V_{\phi}(\mathbf{X}))||m_k)$ works as a class-wise logit.

\vspace{1\baselineskip}
\noindent\textbf{Auxiliary simplex prototypes for training.}\quad
In the post-hoc refinement strategy of Eq.~\eqref{eq:clusterprob}, one can formulate refined class-wise simplex means $\{m_k\}_{k=1}^K$ after finishing the training process of $V_{\phi}$.
In other words, the training process of $V_{\phi}$ is still dependent on the inherent class prototypes of $P_{\theta}$.
To incorporate the prototype refinement principle into the training phase for a further enhancement, we introduce auxiliary simplex prototypes $\mathbf{a} = \{a_k\}_{k=1}^K$.

Initializing the auxiliary prototypes by clustering with initial classification prediction probabilities $\{P_{\theta}(\mathbf{X}_i)\}_{i=1}^N$,
one can update the prototypes throughout the training process,
where the corresponding prediction refinement can be conducted by
\begin{equation}\label{eq:auxprob}
    P_\mathbf{a}(V_{\phi}(\mathbf{X}))[k] = \frac{\exp(-\text{KL}(P_{\theta}(V_{\phi}(\mathbf{X}))||a_k))}{\sum_{l=1}^K \exp(-\text{KL}(P_{\theta}(V_{\phi}(\mathbf{X}))||a_l))}.
\end{equation}

In summary, traditional black-box visual prompting methods manipulate only the input space of PTMs by applying visual prompts to images, limiting learning flexibility due to fixed class prototypes. Instead of directly using the output probability vectors from the PTM, our method treats these vectors as feature vectors and calculates probabilistic distances between auxiliary prototypes to obtain refined class predictions.
In the following section, we introduce our training strategies, including prototype updating rules.

\subsection{Training Probabilistic Clusters}\label{sec3.3}

In the previous subsection, we introduced the concept of auxiliary simplex prototypes. These prototypes enable us to fine-tune the output space of PTMs in conjunction with visual prompting at the input space.
We anticipate that introducing flexibility in the output space may yield synergistic effects during VP training. Consequently, we have carefully designed training strategies that minimize conflicts between tuning in the input and output spaces.
This consideration is crucial because we cannot process through PTMs in a straightforward manner;
alternating updates between the input and output spaces separately may lead to unstable training.

\vspace{1\baselineskip}
\noindent\textbf{Objective function $\mathcal{L}$.}\quad
When tuning a pre-trained visual recognition model,
one can obtain $\{P_{\theta}(V_{\phi}(\mathbf{X}_i))\}_{i=1}^N$ and $\{P_\mathbf{a}(V_{\phi}(\mathbf{X}_i))\}_{i=1}^N$ with the training data $\{(\mathbf{X}_{i}, y_{i})\}_{i=1}^N$, prediction probabilities of the black-box vision-language model and their refined version based on Eq.~\eqref{eq:auxprob}, respectively.
Using the conventional cross-entropy loss function, we define $\mathcal{L}_\text{cls} = \mathcal{L}_\text{CE}(\{P_{\theta}(V_{\phi}(\mathbf{X}_i)), y_i\}_{i=1}^N)$ and $\mathcal{L}_\text{aux} = \mathcal{L}_\text{CE}(\{P_\mathbf{a}(V_{\phi}(\mathbf{X}))\}_{i=1}^N)$ for classification.

In addition to $\mathcal{L}_\text{cls}$ and $\mathcal{L}_\text{aux}$,
we employ an additional loss function to enhance the alignment between the sets $\{P_{\theta}(V_{\phi}(\mathbf{X}_i))\}_{i=1}^N$ and $\{P_\mathbf{a}(V_{\phi}(\mathbf{X}_i))\}_{i=1}^N$.
For this purpose,
we adapt the intra-class relation loss~\cite{huang2022knowledge},
which is proposed to share the relational dynamics
within each class in knowledge distillation.
In our scenario, the relation loss $\mathcal{L}_\text{intra}$
focuses on preserving the relative rankings or preferences of instances within a class,
rather than replicating exact probability values from $\{P_{\theta}(V_{\phi}(\mathbf{X}_i))\}_{i=1}^N$ to $\{P_\mathbf{a}(V_{\phi}(\mathbf{X}_i))\}_{i=1}^N$.
Unlike traditional methods that require strict output matching, $\mathcal{L}_\text{intra}$ adopts a more relaxed approach, prioritizing the correlation of predictions.
In summary, our overall loss function is defined by
\begin{equation}\label{eq:loss}
    \mathcal{L} = \mathcal{L}_\text{cls} + \mathcal{L}_\text{aux} +  \mathcal{L}_\text{intra}.
\end{equation}



\vspace{1\baselineskip}
\noindent\textbf{Updating rules of $\mathbf{a}$.}\quad
Following the zeroth-order optimization method of the recently proposed black-box VP~\cite{blackvip23},
we use the principle of the SPSA-GC algorithm to train our visual prompter $V_{\phi}$.
Instead of updating $\mathbf{a}$ by computing gradients, which may not align with the gradients computed in the SPSA-GC algorithm for input-space VP training, we continuously update auxiliary simplex prototypes following subsequent updates of the visual prompters.
Starting from the initial prototypes $\mathbf{a}_0 = \{a_{0,k}\}_{k=1}^K$, we update the prototypes by selectively obtaining the probabilistic distributions that result in a diminished loss direction. 

To be specific, we discern between the outcomes of bidirectional perturbations ($\phi_t \pm c_t \Delta_{s}$ in Eq.~\eqref{eq:spsagrad}), and incorporate the probabilistic distribution that have lesser loss, thus having superior and accurate probabilistic representation.
Averaging class-wise prediction probabilities,
we refine the prototype vector in a weighted sum manner.
Denoting $p_{t,k}$ as the average of the probability predictions for the $k$-th class training samples at the $t$-th iteration, we compute
\begin{equation}\label{eq:update}
    a_{t+1,k} = \texttt{L1}(0.9 \cdot a_{t,k} + 0.1 \cdot p_{t,k}),
\end{equation}
where $\texttt{L1}$ is $l_1$-normalization to maintain the probabilistic integrity of auxiliary simplex prototypes.


\vspace{1\baselineskip}
\noindent\textbf{Gradient surgery.}\quad
Although SPSA-GC has achieved faster convergence during training,
the black-box optimization algorithm requires
careful hyperparameter tuning to acquire desired performance.
Noting that the multi-point gradient estimation of Eq.~\eqref{eq:spsagrad} simply averages $S$ estimated gradients to build a more stable gradient,
we incorporate the principle of gradient surgery
for robustness.

Inspired by a multi-task learning method which projects a conflicting gradient onto the normal plane of a gradient of other task~\cite{pcgrad20},
we seek to refine the estimation of a conflicting gradient.
Formally, let $\hat{g}_{t_{1}},\hat{g}_{t_{2}}$ be a pair of sub-gradients among the the $S$ gradient estimations obtained concurrently at the same iteration step in Eq.~\eqref{eq:spsagrad}.
Then, the gradient projection process can be formulated by
\begin{equation}
\hat{g}_{t_{1}} = 
\begin{cases} 
\hat{g}_{t_{1}} - \frac{\hat{g}_{t_{1}} \cdot \hat{g}_{t_{2}}}{\|\hat{g}_{t_{2}}\|^2}\hat{g}_{t_{2}} & \text{if } \hat{g}_{t_{1}}, \cdot \hat{g}_{t_{2}} < 0 \\
\hat{g}_{t_{1}}, & \text{otherwise.}
\end{cases}
\end{equation}
We update each gradient estimation relative to other estimations to ensure that the revisions encompass all possible interactions between the gradients under consideration.

\begin{table}[!t]
\begin{center}
\fontsize{8.5}{10pt}\selectfont
\begin{tabular}{c|C{0.4in}C{0.4in}C{0.4in}C{0.4in}}

\toprule

Method & SVHN & Pets & EuroSAT & RESISC \\

\midrule

Spatial (w/ Enc.)    & 58.03 & 89.70 & 73.10 & 64.50 \\
Spatial (w/o Enc.)   & 60.37 & 88.04 & 65.73 & 63.18 \\
Low freq. (w/ Enc.)  & 66.57 & 88.55 & 69.04 & 61.67 \\
Low freq. (w/o Enc.) & 67.35 & 89.45 & 67.69 & 62.31 \\
\bottomrule
\end{tabular}

\end{center}

\caption{
Exploratory results using BlackVIP (Spatial w/ Enc.) and its
variants (Spatial w/o Enc., Low freq. w/ and w/o Enc.).
}

\label{tb:pre_freq}
\end{table}

\section{Experiments}

In this section, we describe the experimental settings (Section~\ref{subsec:setup}) including implementation details, datasets, and baselines.
Before we present our main results, we first introduce exploratory results related to our visual prompter (Section~\ref{sec:exploratory}).
Afterwards, we compare the performance of our model with baseline methods on benchmark and synthetic datasets (Section.~\ref{subsec:results}),
and then show performance improvement by each contribution (Section.~\ref{subsec:ablation}).


\begin{table*}[!ht]
\begin{center}
\fontsize{8}{10pt}\selectfont
\begin{tabular}{c|C{0.29in}C{0.29in}C{0.29in}C{0.29in}C{0.29in}C{0.29in}C{0.29in}C{0.29in}C{0.29in}C{0.29in}C{0.29in}C{0.29in}C{0.29in}C{0.29in}|C{0.18in}}
\toprule

Method & Caltech & Pets & Cars & Flowers & Food & Aircraft & SUN & DTD & SVHN & EuroSAT & RESISC & CLEVR & UCF & IN & Avg. \\

\midrule
\midrule

ZS & 92.9 & 89.1 & 65.2 & 71.3 & 86.1 & 24.8 & 62.6 & 44.7 & 18.1 & 47.9 & 57.8 & 14.5 & 66.8 & 66.7 & 57.6 \\
BAR & 93.8 & 88.6 & 63.0 & 71.2 & 84.5 & 24.5 & 62.4 & 47.0 & 42.7 & 77.2 & 65.3 & 18.7 & 64.2 & 64.6 & 61.4 \\
VP (Black) & 89.4 & 87.1 & 56.6 & 67.0 & 80.4 & 23.8 & 61.2 & 44.5 & 61.3 & 70.9 & 61.3 & 25.8 & 64.6 & 62.3 & 61.2 \\
BlackVIP & 93.7 & \textbf{89.7} & 65.6 & 70.6 & 86.6 & 25.0 & 64.7 & 45.2 & 58.0 & 73.1 & 64.5 & 36.8 & 69.1 & 67.1 & 65.0 \\
\midrule
\textbf{Ours} & \textbf{94.3} & 89.5 & \textbf{69.8} & \textbf{88.8} & \textbf{88.3} & \textbf{33.4} & \textbf{71.5} & \textbf{59.7} & \textbf{73.5} & \textbf{80.3} & \textbf{74.0} & \textbf{45.7} & \textbf{71.0} & \textbf{75.3} & \textbf{72.5} \\
\bottomrule
\end{tabular}
\end{center}

\caption{
The experimental results evaluating the classification performance of our framework and baseline methods in visual recognition tasks across 14 distinct benchmarks, including natural, specialized, structured, and fine-grained categories. Our approach is highlighted for its superior performance among black-box input-space prompting techniques.
The experiments were conducted using a 16-shot learning framework, with each set of experiments repeated three times to ensure reliability.}

\label{tb:main_table}

\end{table*}

\subsection{Experimental Settings}
\label{subsec:setup}

\textbf{PTMs and Baselines.}\quad
As a representative visual recognition PTM, we employed the CLIP zero-shot classifier~\cite{radford2021learning}, which can be extended across various numbers of classes and multi-modal scenarios.
Specifically, along with each image $\mathbf{X}$,
CLIP typically takes a set of text prompts based on a hand-crafted template (\emph{e.g.}, ``\texttt{a photo of a [CLASS].}'') and then extract text features via its tokenizer and text encoder.
These extracted features are instrumental in generating class prediction probabilities.

For the baseline methods,
we considered CLIP zero-shot classification (ZS), BAR~\cite{reprogramming20},
a black-box version of the visual prompting method introduced in~\cite{bahng2020learning}, and BlackVIP~\cite{blackvip23}.

\vspace{1\baselineskip}
\noindent\textbf{Implementation details.}\quad
As aforementioned, we utilized CLIP VIT-B/16~\cite{radford2021learning} for a frozen PTM $P_\theta$, which is capable of strong zero-shot generalization.
We adopted the decoder design of BlackVIP~\cite{blackvip23}, in which low-frequency visual prompts (VPs) of size ($56 \times 56 \times 3$) were generated by applying a $1 \times 1$ convolution operation to its intermediate layer.
As the default input size of $P_\theta$ was ($224 \times 224 \times 3$), we resized each image into ($112 \times 112 \times 3$) or ($192 \times 192 \times 3$) to apply visual prompts, according to its original resolution.
For the scaling parameter $\phi^s$, we initialized it at $\phi^s = -5$ so that $\sigma(\phi^s) \approx 0$, thereby minimizing the impact of frequency-domain prompting in the early training stages.
Additionally, this parameter was designed to be adjustable throughout training.



\vspace{1\baselineskip}
\noindent\textbf{Datasets.}\quad
Following the protocol of BlackVIP~\cite{blackvip23},
we used 14 image classification benchmark datasets and few-shot training \& validation: 16-shot for training and 4-shot for validation.
The datasets include a wide range of domains such as natural image, remote sensing, and textures, which can evaluate our framework on diverse modalities.
For further analyses,
we additionally employed two synthetic datasets: Biased MNIST and Loc-MNIST.

\subsection{Exploratory Results}
\label{sec:exploratory}



We first outline exploratory results that underpin the development of our visual prompter design.
These observations are essential for guiding the architectural decisions and functional strategies.

Recall the framework of BlackVIP~\cite{blackvip23}, which makes spatial-domain VPs of the same size as each image based on an encoder-decoder structure.
To design an effective visual prompter without encoder,
we first constructed three variants of BlackVIP (spatial-domain VP with an encoder) and test them with a number of datasets, SVHN, Oxford-Pets, EuroSAT, and RESISC.
Here, the variants are 1) spatial-domain VP without encoder, 2) low-frequency VP with an encoder, and 3) low-frequency VP without encoder, where we followed the experimantal settings of BlackVIP.
In Fig.~\ref{fig:visualprompting}, we illustrate various visual prompting methods for a better understanding of spatial- and frequency-domain prompting.



Table~\ref{tb:pre_freq} illustrates that frequency-domain prompting can successfully adapt to certain datasets, such as SVHN, although it may yield sub-optimal results in other contexts.
To leverage the benefits of both spatial and frequency domain prompting without the drawbacks of each interfering, we have developed our visual prompter as shown in Fig.~\ref{fig:overview}.
In this design, we introduce a learnable scaling parameter, $\phi^s$, which enables $V_\phi$ to dynamically adjust the impact of low-frequency prompting. This allows for customized application, tailored to maximize effectiveness in various contexts.

We evaluated the impact of incorporating an additional trigger vector in dual-domain visual prompt generation by comparing training outcomes with and without it. Anticipating that the trigger vector would provide increased learning flexibility and enhance adaptability across different domains, our findings confirmed that its inclusion significantly enhanced both performance and the overall learning process compared to the version without it.

\begin{table*}[!ht]
\begin{center}
\fontsize{8.5}{10pt}\selectfont
\begin{tabular}{c|C{0.4in}C{0.4in}|C{0.4in}C{0.4in}|C{0.4in}C{0.4in}|C{0.4in}C{0.4in}}

\toprule

\multirow{3.2}{*}{Methods} & \multicolumn{4}{c|}{Biased MNIST} & \multicolumn{4}{c}{Loc-MNIST}\\

\cmidrule{2-9}

 & \multicolumn{2}{c|}{16-shot} & \multicolumn{2}{c|}{32-shot} & \multicolumn{2}{c|}{16-shot} & \multicolumn{2}{c}{32-shot} \\

\cmidrule{2-9}

 & $\rho = 0.8$ & $\rho = 0.9$ & $\rho = 0.8$ & $\rho = 0.9$ & $1:1$ & $1:4$ & $1:1$ & $1:4$ \\

\midrule
\midrule

ZS & 37.56 & 37.25 & 37.56 & 37.25 & 29.70 & 22.70 & 29.70 & 22.70 \\
BAR & 53.25 & 53.07 & 53.93 & 53.30 & 33.98 & 26.05 & 34.73 & 27.72 \\
VP (Black) & 60.34 & 53.86 & 59.58 & 51.88 & 16.21 & 25.68 & 18.43 & 30.13 \\
BlackVIP & 66.21 & 62.47 & 65.19 & 64.47 & \textbf{69.08} & 60.86 & 76.97 & 67.97 \\
\midrule
\textbf{Ours} & \textbf{73.17} & \textbf{75.16} & \textbf{76.53} & \textbf{76.38} & 69.01 & \textbf{65.16} & \textbf{77.10} & \textbf{70.31} \\
\bottomrule
\end{tabular}

\end{center}

\caption{
The experimental results evaluating the robustness against distribution shift of our framework and baseline methods.
Using two synthetic datasets, Biased MNIST and Loc-MNIST,
the experiments were conducted using 16- and 32-shot learning frameworks, with each set of experiments repeated three times to ensure reliability.}
\label{tb:synthetic}
\end{table*}

\begin{table}[!t]
\begin{center}
\fontsize{8.5}{10pt}\selectfont
\begin{tabular}{cl|C{0.37in}C{0.37in}C{0.37in}}

\toprule

\multicolumn{2}{c|}{Components} & SVHN & EuroSAT & DTD \\

\midrule

(1) & Hybrid visual prompting        & 70.1 & 75.0 & 48.5 \\
(2) & + Prediction refinement        & 71.3 & 76.7 & 51.1 \\
(3) & + Auxiliary simplex training   & 72.4 & 78.9 & 58.0 \\
(4) & + Intra-class relation loss    & 73.0 & 79.6 & 58.9 \\
(5) & + Gradient surgery             & 73.5 & 80.3 & 59.7 \\
\bottomrule
\end{tabular}
\end{center}

\caption{
Ablation study results, where we incrementally integrated the proposed techniques: spatial-frequency visual prompting, prediction refinement, auxiliary simplex training, intra-class relation loss, and gradient surgery.}
\label{tb:ablation}
\end{table}

\subsection{Experimental results}
\label{subsec:results}

\textbf{Transfer learning on benchmark datasets.}\quad
Utilizing 14 benchmark datasets, we implemented a few-shot approach as described in previous studies~\cite{exploring22, zhou2022conditional, zhou2022learning} to conduct our evaluations, with the results detailed in Table~\ref{tb:main_table}.
The findings suggest that our method consistently outperforms existing black-box visual prompting techniques in terms of overall performance. 



Our method has demonstrated considerable superiority across multiple datasets by substantial margins, underscoring the effectiveness of our visual prompter.
This enhanced performance is particularly evident when compared to various baseline variants, as detailed in Table~\ref{tb:pre_freq}.
These comparisons highlight the benefits of our hybrid approach, illustrating the potential for improved adaptation by synergistically utilizing both spatial and frequency information, thereby validating the design's efficacy in boosting model performance across diverse testing environments.

\vspace{1\baselineskip}
\noindent\textbf{Transfer learning on synthetic datasets.}\quad
To evaluate the robustness of visual prompting methods against distribution shifts, we followed the experimental protocol outlined in BlackVIP~\cite{blackvip23}, using datasets such as Biased MNIST~\cite{bahng2020learning} and Loc-MNIST. This approach allows us to systematically assess the effectiveness of these methods in varying data environments.

Biased MNIST is designed to examine generalization ability under color bias shift.
At training, every digit was assigned a distinct background color that is closely linked to its class label, where the correlation strength is set by the value of $\rho \in [0,1]$.
In test phase, the correlation ratio was inverted to $1-\rho$. Table~\ref{tb:synthetic} shows that our approach yields better outcomes than the previous methods.
These results may indicate that input-dependent prompting isn't always necessary for distribution shift robustness, as it can lead to adverse effects when unrelated information is added to images.

To further validation,
Oh~\emph{et al.}~\cite{blackvip23} has developed 
Loc-MNIST, which involves placing an actual target digit on one of the four edges and a random false digit in the middle of a blank image.
Both the position of the target digit and the type of the fake digit are randomly selected.
Additionally, a more complex scenario where the fake digit is four times bigger (1:4 ratio) than the real one is considered.
We also present the corresponding results in Table~\ref{tb:synthetic},
which complements the analysis using Biased MNIST.



\subsection{Ablation Study and Analysis}
\label{subsec:ablation}


\noindent\textbf{Ablation study.}\quad
To assess the individual contributions of the components in our black-box visual prompting method, we conducted an ablation study.
Beginning with our spatial-frequency hybrid visual prompting, we gradually incorporated the proposed training and inference techniques: prediction refinement, auxiliary simplex training, intra-class relation loss, and gradient surgery.

Referring to Table~\ref{tb:main_table}, it is evident from Table~\ref{tb:ablation} that our hybrid visual prompter achieves higher classification accuracy compared to existing state-of-the-art black-box tuning methods. Additionally, it is clear that employing prediction refinement through KL k-means clustering consistently enhances recognition accuracy. The auxiliary simplex training strategy further improves accuracy, particularly in certain datasets (\emph{e.g.}, DTD). Furthermore, the intra-class relation loss and gradient surgery strategies have proven effective in stabilizing the training process.

\begin{figure}[t] 
\begin{center} 
\includegraphics[width=0.95\linewidth]{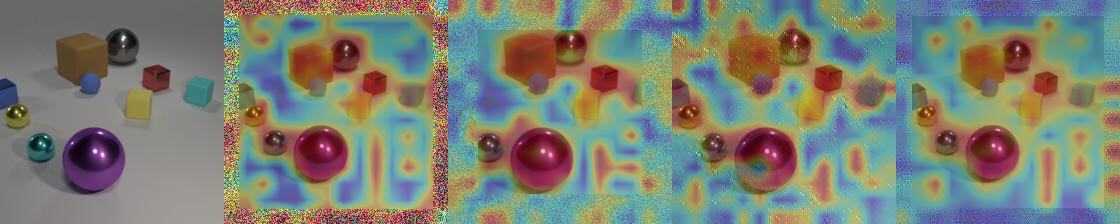}\\
\includegraphics[width=0.95\linewidth]{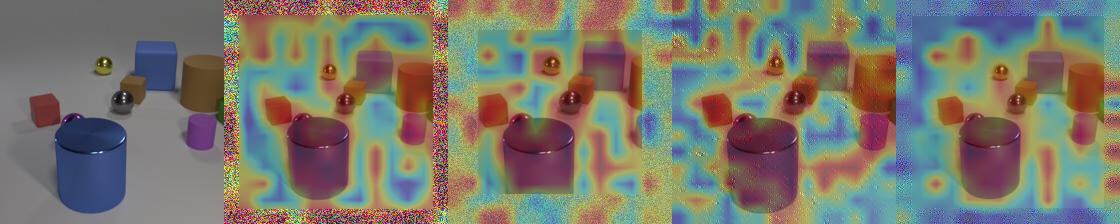}
\end{center}
\caption{From left to right, we present images sampled from the CLEVR dataset and the corresponding GradCAM analysis results of BAR, VP (Black), BlackVIP, and our method.}
\label{fig:cam_results}
\end{figure}

\vspace{1\baselineskip}
\noindent\textbf{Visual analysis.}\quad
To further investigate the effectiveness of our propose method, we provide visual demonstration by overlaying GradCAM~\cite{selvaraju2017grad} results on the visual-prompted images sampled from the CLEVR dataset.
From left to right, Figure~\ref{fig:cam_results} presents original images and the corresponding GradCAM visualization results of BAR, VP (black-box), BlackVIP, and ours, respectively.

Note that the visual recognition task in CLEVR involves counting the number of objects in each image. Thus, it is crucial to accurately identify the exact location of each object. As shown in the figure, our method outperformed the other baselines in this aspect.



\vspace{1\baselineskip}
\noindent\textbf{Computational costs.}\quad
In contrast to BlackVIP~\cite{blackvip23}, a state-of-the-art black-box visual prompting technique that uses an encoder-decoder architecture, our approach achieves greater efficiency during training, due to the omission of the self-supervised encoder.

Through experimental evaluation on the SVHN and EuroSAT datasets, using an NVIDIA RTX A5000 GPU, our method demonstrated a more resource-efficient footprint, consuming 2971MB of GPU memory and requiring only 2.24 seconds per iteration. In comparison, the same metrics for other baselines were as follows: BAR (4049MB, 2.80 sec/iter), VP (4065MB, 4.32 sec/iter), and BlackVIP (4223MB, 3.75 sec/iter). These results highlight the efficiency gains of our methodology, underscoring its practical advantages.

\section{Conclusion}

In conclusion, our work has successfully implemented a novel transfer learning framework for large PTMs that effectively bridges the data distribution gap in resource-constrained black-box API environments. By integrating spatial-frequency hybrid visual prompts with cluster-based prediction refinement, we have achieved significant improvements in few-shot transfer learning performance across a variety of datasets. This approach also results in reduced computational costs. Our methods enhance the adaptability and efficiency of PTMs in practical applications, while also fostering new research opportunities in transfer learning and model adaptation. The progress documented in this paper sets the stage for further investigation and advancement in the field, promising to enhance the utility and reach of machine learning across multiple scenarios.



\begin{acks}
This work was supported by Institute for Information \& communications Technology Promotion (IITP) grant funded by the Korea government (MSIT) (No. RS-2019-II190075 Artificial Intelligence Graduate School Program (KAIST)), the National Research Foundation of Korea (NRF) grant funded by the Korea government (MSIT)(No. 2022R1A5A7083908), and the Starting growth Technological R\&D Program (No. RS-2023-00272605) funded by the Ministry of SMEs and Startups (MSS, Korea).
\end{acks}

\bibliographystyle{ACM-Reference-Format}
\balance
\bibliography{citation}

\end{document}